\documentclass{vgtc}                          

\graphicspath{{figures/}{pictures/}{images/}{./}} 

\usepackage{times}                     

\usepackage{placeins}

\usepackage{mathptmx}                  
\usepackage{amsmath}
\usepackage{array}

\onlineid{0}

\vgtccategory{Research}

\vgtcinsertpkg
\usepackage{dblfloatfix}
\usepackage{float}

\title{Attention Dynamics in Diffusion Models: A Visual Analytics Framework for Human--AI Collaboration}

\author{Yiran Xiao\thanks{e-mail: yiranxiao@ucsb.edu} %
\and George Legrady\thanks{e-mail: glegrady@ucsb.edu}}
\affiliation{\scriptsize University of California, Santa Barbara\\Santa Barbara, California, United States}

\teaser{
  \centering
  \makebox[\textwidth][c]{\includegraphics[width=1.18\textwidth]{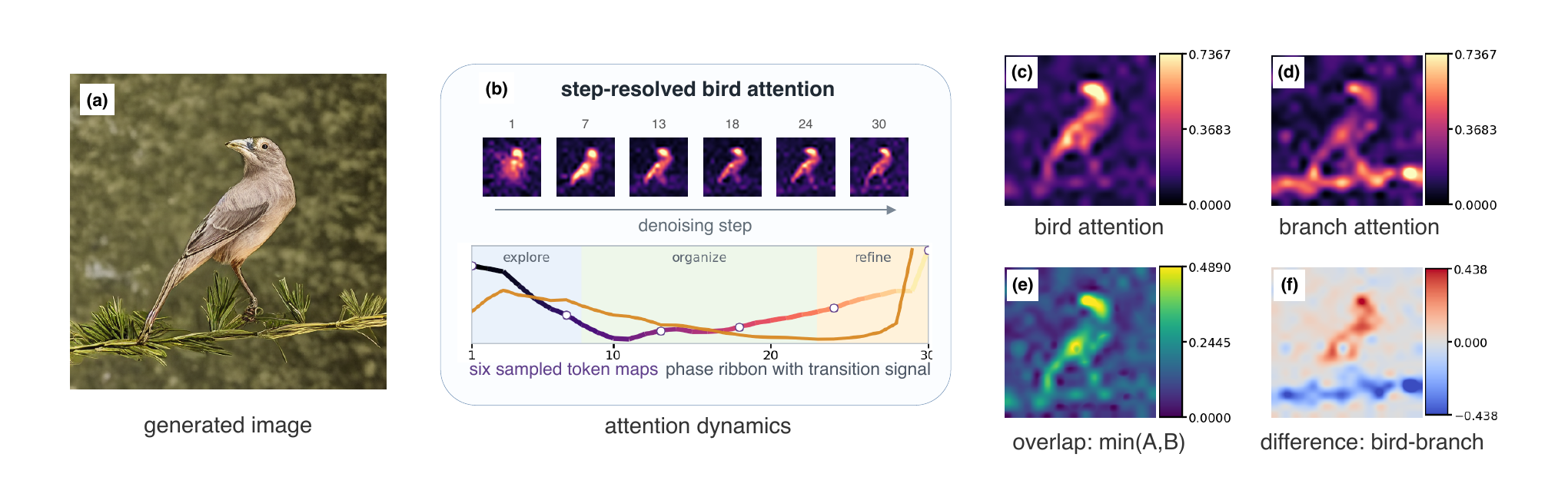}}
  \caption{\textbf{Overview of step-resolved attention analysis.} For \texttt{bird\_04} (\emph{a bird on a branch}), the teaser connects the synthesized image to the temporal and spatial evidence used throughout the system. Panel (b) samples \textsf{bird} attention across denoising steps and summarizes the trajectory with a phase-aware temporal ribbon, highlighting how semantic localization stabilizes over generation. Panels (c--f) show the late-step token-competition view: individual \textsf{bird} and \textsf{branch} maps, their shared support $\min(A,B)$, and the signed spatial difference \textsf{bird}$-$\textsf{branch}.}
  \label{fig:teaser}
}

\abstract{Diffusion-based text-to-image models can synthesize complex and highly structured visual content, yet the emergence and evolution of semantic structure remain difficult to interpret. Many existing workflows rely on aggregated attention or scalar summaries that separate temporal change from image-space evidence. To address this gap, we present a visual analytics framework for exploring \emph{attention dynamics} in diffusion models: the step-indexed evolution of token-level cross-attention maps, their temporal concentration, and their spatial relationships. Our approach enables structured analysis of attention behavior across generation steps by integrating quantitative measures with data-driven stage identification in an interactive workflow. Case studies on a structured 60-prompt Stable-Diffusion-class benchmark illustrate recurring, interpretable patterns within this setting and show how linked temporal and spatial views facilitate the observation and discussion of generative processes, supporting more effective human--AI collaboration.
} 


\begin{document}


\firstsection{Introduction}

\maketitle

\begin{figure*}[t]
  \centering
  \includegraphics[width=0.98\textwidth]{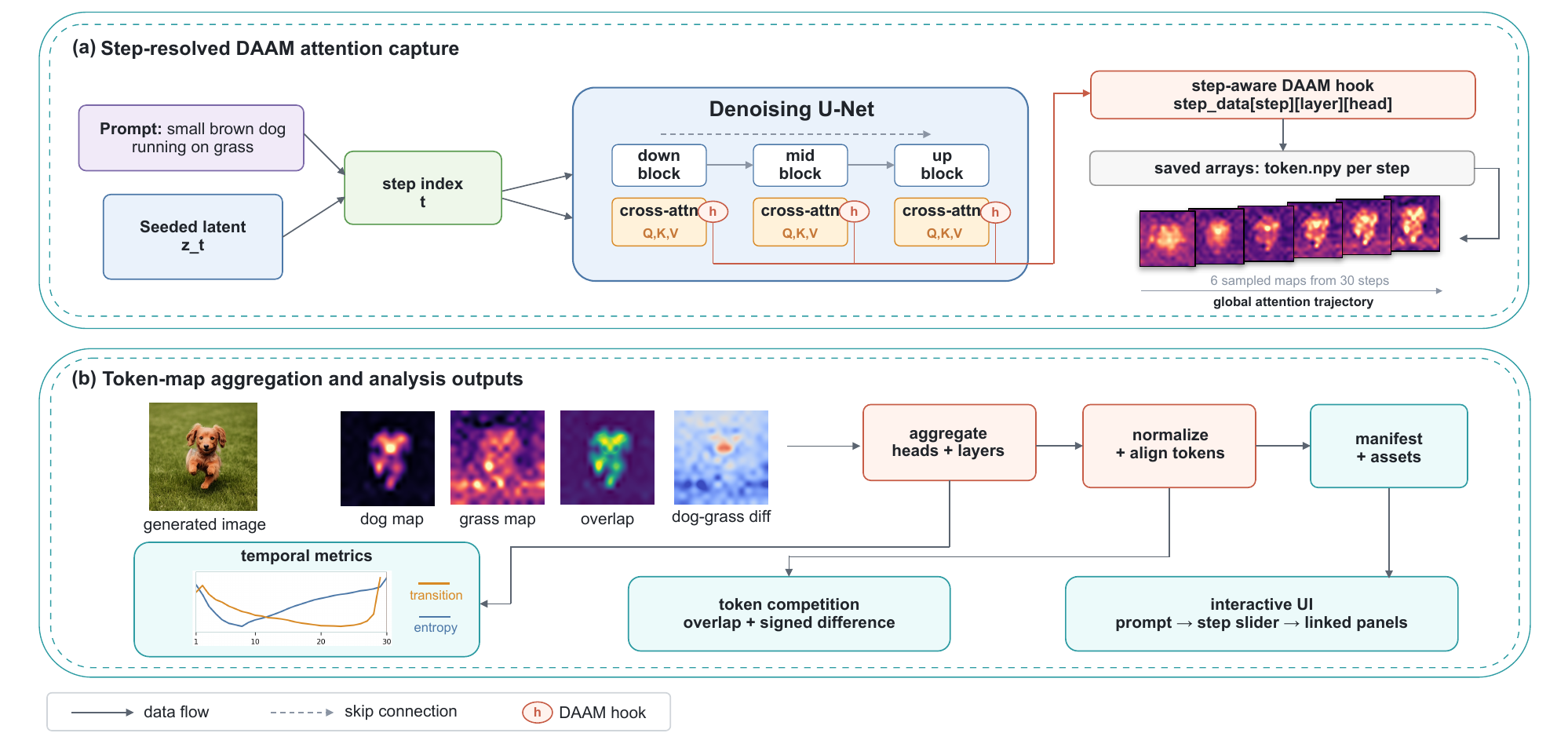}
  \caption{\textbf{DAAM-style attention capture pipeline.} DAAM (Diffusion Attentive Attribution Maps) estimates word-level diffusion attribution from cross-attention~\cite{tang2023daam}; our step-resolved capture records heatmaps by denoising step, U-Net layer, and attention head. (a) The stack shows six representative global maps sampled from the full 30-step schedule. (b) Captured maps are aggregated and normalized into aligned token maps $A_t^{token}$, supporting temporal metrics, token-competition views, and the interactive analysis interface.}
  \label{fig:daam_pipeline}
\end{figure*}

Text-to-image diffusion models can synthesize coherent scenes from short prompts~\cite{ho2020ddpm,song2021ddim,rombach2022high,zhang2023t2isurvey,wang2025diffusionart}, yet the process by which semantic structure emerges remains difficult to inspect. This opacity matters for both model developers and downstream creators: a generated image may look plausible, but the user still has little evidence about when the model localized an object, when an attribute became attached to it, or whether two prompt tokens competed for the same spatial region. Existing interfaces therefore leave analysts with a familiar gap between impressive outputs and weak process-level explanations.

Cross-attention offers a useful window into this gap because each denoising step relates text tokens to spatial latent positions~\cite{hertz2022prompt,tang2023daam,hua2025attentiondiffusion}. In this paper, we define \emph{attention dynamics} as the step-indexed evolution of token-level cross-attention maps during denoising, including changes in concentration, step-to-step movement, and spatial overlap or dominance between prompt tokens. Common inspection workflows either aggregate attention into a final word-level heatmap or show many per-token curves without the image-space evidence needed to interpret them. Final maps hide temporal reorganization, while scalar curves hide where the corresponding attention mass moves. For visual analytics, the important unit is not only a token map, but a linked trajectory: \emph{when} attention changes, \emph{where} it changes, and \emph{which} prompt tokens share or separate space.

We present a visual analytics framework for step-resolved diffusion attention. The system captures token maps using Diffusion Attentive Attribution Maps (DAAM) cross-attention attribution~\cite{tang2023daam}, organizes prompts into controlled semantic families, and links temporal summaries with spatial token-competition panels in an interactive UI. Users locate broad phases in a global timeline, move a step cursor to inspect maps, and compare token pairs through overlap and signed-difference views. This workflow complements final-map attribution and pipeline explanation systems by asking when attention changes, where the change occurs, and which prompt tokens share or separate image space.

Our contributions are: \textbf{(1)} a DAAM-based, step-resolved capture and data-organization pipeline for token-level attention trajectories; \textbf{(2)} a visual analytics workflow linking established temporal and spatial summaries--concentration, transition, overlap, and local dominance--across controlled prompt families; and \textbf{(3)} case studies and pilot evidence showing how linked phase and token competition views help users inspect diffusion attention dynamics beyond final maps or scalar curves alone.

\section{Related Work}
\label{sec:related_work}
\textbf{Diffusion attention and attention maps.}
Latent diffusion models generate images by iteratively denoising a latent representation while conditioning the U-Net on text embeddings~\cite{rombach2022high}. In cross-attention layers, each text token contributes attention weights over spatial latent positions. Aggregating these weights for a word yields an \emph{attention map}: a spatial heatmap that indicates where the model allocates token-related attention at a particular layer, head, or denoising step. Attention maps are not ground-truth object masks, but they provide a useful attribution signal for inspecting how text conditioning is distributed across the image~\cite{park2024explaining,liu2024crossselfattention}.

Cross-attention has become a common mechanism for editing and interpretation. Prompt-to-Prompt modifies cross-attention maps to preserve layout while changing text~\cite{hertz2022prompt}, and DAAM aggregates diffusion attention to estimate word-level image attribution~\cite{tang2023daam}. Attention-based guidance and grounding methods further use attention to improve semantic faithfulness, compositional binding, phrase localization, and visual perception~\cite{chefer2023attend,feng2023structured,wang2024attentioncontrol,zhao2023vpd,liu2024vgdiffzero,yang2024diffpng,zhan2023stable3d}; recent surveys document growing interest in attention behavior across diffusion models~\cite{hua2025attentiondiffusion}. Our work does not replace these methods: it uses DAAM as the attribution baseline and contributes a step-indexed visual organization for comparing how token maps change and how semantic token pairs share or separate image space.

\textbf{Visual analytics for model behavior.}
Visualization research externalizes intermediate model states so analysts can compare and communicate patterns rather than inspect raw logs~\cite{heer2012interactive,sedlmair2012design}. For text-to-image systems, Diffusion Explainer visualizes Stable Diffusion mechanics for interactive learning~\cite{lee2024diffusionexplainer}, while PrompTHis and PromptNavi study prompt editing and visual exploration~\cite{guo2024prompthis,huang2025promptnavi}. Prior HCI and XAI work further argues that generated-image interpretations should connect back to visual evidence~\cite{rapp2025genaiimages,nuthalapati2025reversemapping}. Our narrower scope is to connect DAAM-style token-attention traces to spatial maps so readers can verify a curve against image-space evidence at the same denoising step.

Mechanistic approaches such as Revelio learn sparse autoencoder features inside diffusion models~\cite{kim2024revelio}. We instead organize already-captured cross-attention maps into inspectable evidence. The distinction is one of workflow and evidence structure: DAAM provides word-level attribution, Diffusion Explainer teaches the diffusion process, and our system supports step-resolved comparison of token trajectories, phase summaries, and token-pair spatial competition over a controlled prompt benchmark.

\section{System Overview}
\label{sec:system}
Our workflow, summarized in \cref{fig:daam_pipeline}, computes attention evidence offline from a frozen Stable Diffusion v1.5 pipeline (\texttt{runwayml/stable-diffusion-v1-5}) using the Euler scheduler, 30 denoising steps, guidance scale 7.0, and $512{\times}512$ outputs on an NVIDIA RTX 4090 workstation. We do not train or fine-tune the model; DAAM serves as the cross-attention attribution baseline, and our changes are step-resolved capture, benchmark organization, analysis measures, and the linked UI. During generation, we hook the DAAM heat-map update routine and record cross-attention heatmaps by denoising step, U-Net layer, and attention head. We then aggregate the maps across layers and heads, align them to selected prompt tokens, and normalize them into step-indexed token maps $A_t^{token}$. For each prompt, we use 16 shared seeds: all tokens and token pairs are evaluated under the same noise conditions, treating seed as a random factor rather than trying to remove it. This paired design reduces variance in token comparisons because seed choice can strongly affect text-to-image outputs~\cite{xu2024secretseeds}.

\begin{table}[t]
  \centering
  \small
  \setlength{\tabcolsep}{3pt}
  \begin{tabular}{@{}p{0.36\columnwidth}p{0.56\columnwidth}@{}}
    \hline
    \textbf{Prompt family} & \textbf{Comparison emphasized} \\
    \hline
    object only & baseline timing for one semantic token \\
    object--attribute & binding between object and adjective \\
    object--background & foreground/support partition \\
    object--object & direct instance competition \\
    full template & multiple roles in one sentence \\
    \hline
  \end{tabular}
  \vspace{2pt}
  \caption{\textbf{Benchmark organization.} The 60 prompts are structured so each view has a known semantic comparison, rather than relying on arbitrary prompt text.}
  \label{tab:prompt_schema}
\end{table}

\Cref{tab:prompt_schema} summarizes the prompt suite. The benchmark contains 60 structured prompts that add controlled semantic factors: object-only prompts establish baseline timing; object--attribute prompts test binding; object--background prompts expose foreground/support partitioning; and object--object prompts introduce direct competition. The count is not a claim of exhaustive coverage; it provides repeated examples within each semantic family so case-study observations are grounded in controlled comparisons rather than isolated hand-picked prompts. Each prompt record stores semantic components, analysis tokens, token pairs, shared seeds, and paths to cached arrays and figures. This structure keeps the study interpretable and responds to known prompt sensitivity in text-to-image systems and prompt datasets~\cite{wang2022diffusiondb,mo2024dynamicprompt,wang2023tokencompose}.

\subsection{Interface Workflow}
The interface follows a three-step reading workflow. First, the participant selects a prompt family and prompt ID; the global timeline updates to show entropy, transition signal, and phase boundaries. Second, the participant moves the denoising-step slider; a red cursor aligns the timeline with the static panels below so that numerical changes can be checked against maps at the same step. Third, if a token pair is available, the participant compares the two maps, the overlap map, and the signed difference map. This mirrors overview--detail interaction: the timeline provides temporal context, while the spatial strip provides local evidence for claims about binding, support, and competition.

We intentionally avoid live generation in the study UI. Precomputation removes latency, keeps every participant on the same examples, and prevents incidental prompt-engineering differences from becoming the object of the study. The cost is that users cannot freely edit prompts; we treat this as appropriate for the prototype because the central question is whether fixed attention dynamics can be communicated clearly.

\section{Analysis Measures}
\label{sec:analysis}
\textbf{Temporal concentration.}
For each analysis token $w$, seed $s$, and denoising step $t$, we treat the captured map as a nonnegative spatial field and normalize it over pixels: $p_{t,s}^w(i)=A_{t,s}^w(i)/\sum_j A_{t,s}^w(j)$. We then compute Shannon entropy $H_{t,s}^w=-\sum_i p_{t,s}^w(i)\log p_{t,s}^w(i)$ in nats and summarize each token by its mean trajectory over the 16 shared seeds. High entropy indicates that attention is spread over many latent positions; lower entropy indicates a more concentrated token map. For token-pair comparisons, we compute paired per-seed differences, e.g., $\Delta_{t,s}=H_{t,s}^{A}-H_{t,s}^{B}$, before averaging so that common seed effects cancel. Entropy is therefore a temporal index that tells readers where to inspect the corresponding spatial maps.

\textbf{Temporal transition.}
Entropy describes how concentrated a map is, but it does not say whether the map has moved. We therefore compute step-to-step change between consecutive maps. For aligned maps $A_t^w$ and $A_{t+1}^w$, the preprocessing records L2 distance $\lVert A_{t+1}^w-A_t^w\rVert_2$, cosine distance $1-\cos(A_t^w,A_{t+1}^w)$, and L1 change $\sum_i |A_{t+1}^w(i)-A_t^w(i)|$. The global transition curve shown in the timeline averages this evidence across analysis tokens and seeds. Peaks indicate moments when the attention field reorganizes; flat regions indicate intervals where the attention layout is comparatively stable. We segment the denoising schedule into $k{=}4$ named phases to provide a coarse vocabulary for early exploration, semantic anchoring, relational organization, and detail refinement. These labels summarize patterns observed in our current Stable-Diffusion-class benchmark; they are an interface convention, not a claim that diffusion models possess universal cognitive stages.

\textbf{Spatial competition.}
Benchmark metadata defines token pairs such as object--attribute, object--background, and object--object. For a selected pair $(A,B)$, we report a 2$\times$2 spatial strip: token map $A$, token map $B$, pointwise overlap $\min(A,B)$, and signed difference $A-B$. The overlap panel marks shared territory where both tokens assign attention mass; the signed difference panel acts as a local dominance map, with positive regions favoring token $A$ and negative regions favoring token $B$. These panels are intentionally visual rather than scalar: object--attribute pairs often overlap because the attribute is bound to the object, while object--background pairs often separate into foreground and support regions. Spatial strips use the closest saved step to the slider position and are generated from the same paired-seed pipeline.

This split between curve evidence and spatial explanation is important. Entropy is compact and comparable across prompts, but it is abstract; a participant cannot easily infer what an entropy rise looks like in the image. The spatial strip answers that question by showing where mass is assigned. We therefore treat entropy as a temporal index and attention maps as the explanatory view.

\section{Case Studies and Findings}
\label{sec:findings}
\begin{figure}[t]
  \centering
  \includegraphics[width=0.94\columnwidth]{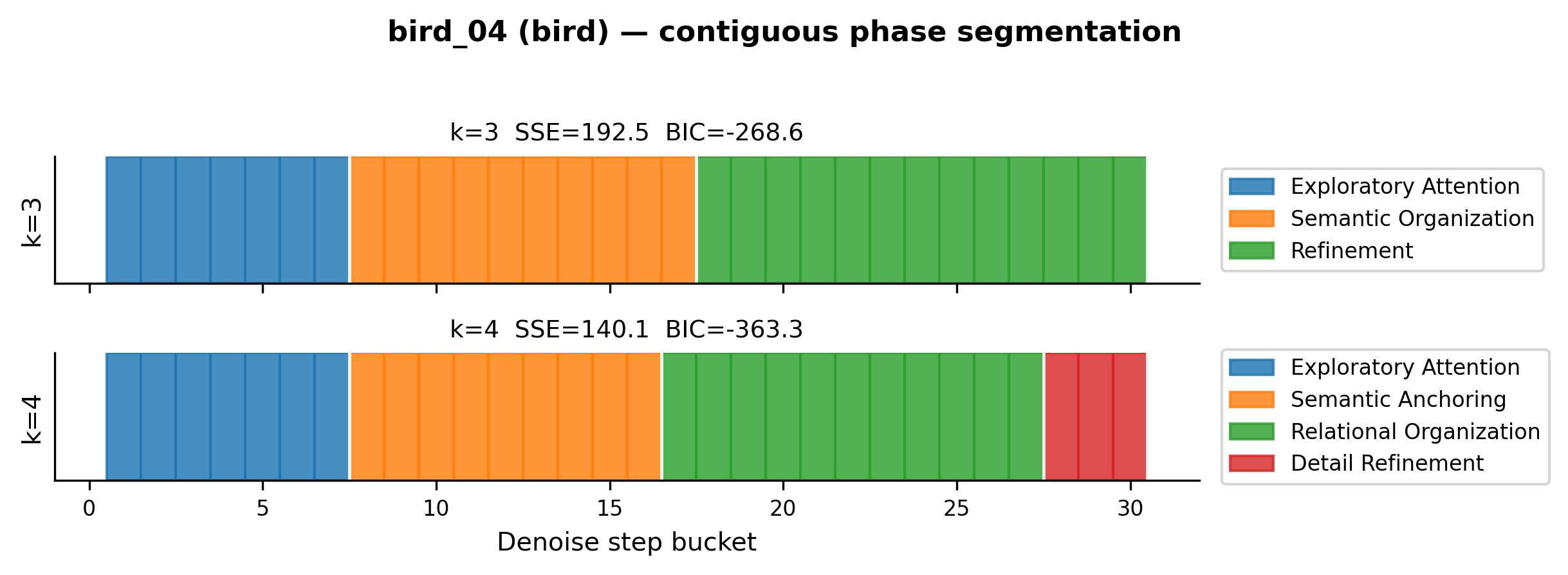}
  \caption{\textbf{Phase-centric summary for \texttt{bird\_04}.} The UI reduces many per-token curves to a global timeline with four named phase bands. This is the main timing view; dense entropy grids are kept as supplemental evidence rather than the first-read figure.}
  \label{fig:findings}
\end{figure}

\textbf{Finding 1: the interface makes coarse-to-fine timing inspectable.}
Across the structured prompt suite and model setting used in this paper, the most useful signal is not the exact shape of every entropy curve, but the ability to locate expected coarse-to-fine behavior in time and then inspect the corresponding maps. \Cref{fig:findings} illustrates the timing view for \texttt{bird\_04}. In the early phase, the entropy and transition curves indicate broad allocation and larger changes between adjacent steps; this is the interval where sampled token maps are still reorganizing. In the middle phase, transition magnitude decreases and token maps become easier to compare spatially. In the late phase, small transition values suggest refinement rather than wholesale relocation. The contribution here is not that coarse-to-fine behavior is surprising, but that phase bands and linked maps make the timing and spatial evidence checkable without reading dense attention arrays.

\begin{figure}[t]
  \centering
  \includegraphics[width=0.94\columnwidth]{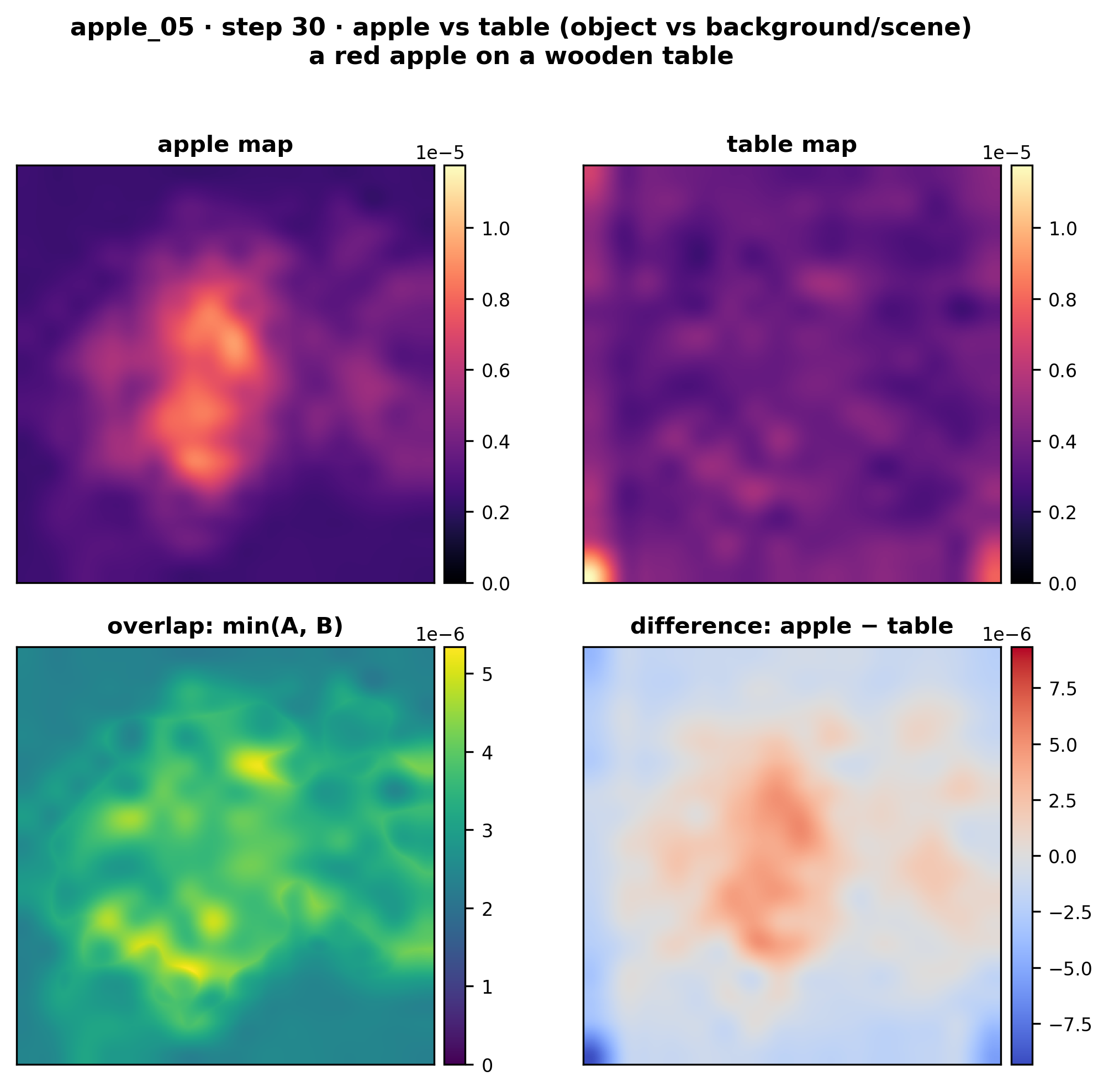}
  \caption{\textbf{Object vs.\ support in space.} For \texttt{apple\_05} (\emph{a red apple on a wooden table}), the late-step strip averages maps over 16 shared seeds and shows \textsf{apple} concentrating on the object, \textsf{table} spreading across support, their overlap, and the signed difference.}
  \label{fig:case_spatial_table}
\end{figure}

\textbf{Finding 2: spatial strips make scalar trends interpretable.}
Token pairs reveal relationships that scalar curves or final maps alone can hide. In \cref{fig:case_spatial_table}, \textsf{apple} remains localized while \textsf{table} spreads across the support surface. The overlap panel highlights regions where the object and support both receive attention, while the signed difference panel recovers the foreground object as a positive region against the broader table field. This makes the scalar story visible: even if entropy reports concentration or diffusion, the spatial strip shows what that concentration means in image space. The teaser gives the same reading for \texttt{bird\_04}: \textsf{bird} and \textsf{branch} can be compared as foreground and support without asking readers to infer spatial behavior from raw entropy lines.

\textbf{Finding 3: prompt families expose different competition patterns.}
The same views support comparisons across prompt families. Object--attribute prompts often show stronger map overlap because attribute attention remains bound to the object; for example, \textsf{red} is expected to share territory with \textsf{apple}. Object--background prompts show clearer spatial separation because support objects such as \textsf{table} or \textsf{branch} occupy broader regions. Object--object prompts can alternate local dominance because both tokens correspond to plausible foreground instances. These patterns are consistent with prior expectations, but the controlled family structure makes them comparable under shared seeds and provides a clearer baseline for noticing departures from the expected pattern.

\FloatBarrier

\section{User Study}
\label{sec:user_eval}
We conducted a lightweight remote pilot study to evaluate whether the interface communicates diffusion attention dynamics to readers with different technical backgrounds. Participants read a plain-language task sheet defining prompts, tokens, denoising steps, stability, competition, and phase segmentation, then used the fixed offline interface so all responses referred to the same prompt IDs, seeds, token pairs, and cached maps.

The study contained four analysis tasks: participants compared stabilization for \textsf{apple} and \textsf{red}, described late-step dominance for an \textsf{apple}--\textsf{red} pair, and reported the number of phases and the largest \textsf{bird}--\textsf{branch} difference for \texttt{bird\_04}. The form also collected six 1--5 usefulness ratings plus open-ended comments.

Pilot responses ($n{=}8$) came from participants with backgrounds in machine learning, visualization/HCI, and text-to-image diffusion. Ratings were positive overall: the entropy/timeline view, phase bar, and process-understanding item each had median ratings of 4 or higher. Most participants identified the four-phase structure for \texttt{bird\_04}, and free-text answers often located the largest \textsf{bird}--\textsf{branch} difference in early-to-middle denoising.

Open-ended feedback highlighted the global timeline, entropy curve, stability-step numbers, and spatial side-by-side maps because they made stepwise attention change, stable moments, and token dominance visible. Several participants used temporal and spatial evidence together rather than relying on final maps or curves alone.

The pilot also revealed design opportunities: clearer phase terminology, stronger slider-map links, hover tooltips, short automatic summaries, and time-varying overlap or dominance curves. One response noted that a missing token-pair asset could load the wrong spatial panel; we fixed this lookup. We treat these results as formative evidence rather than a controlled statistical evaluation.

\section{Limitations and Future Work}
\label{sec:discussion}
The system is designed for readable analysis rather than exhaustive explanation. Attention is an attribution signal, not a complete mechanistic account or a ground-truth mask. Phase labels are reading scaffolds rather than discovered cognitive states, and the $k{=}4$ segmentation is chosen for interface consistency. The benchmark is structured and English-only, and all experiments use one Stable-Diffusion-class backbone and scheduler. We therefore do not claim that the four-phase pattern generalizes to SDXL, DiT/MMDiT-based models such as Flux or Stable Diffusion 3, or alternative samplers such as DDIM and DPM++~\cite{podell2024sdxl,peebles2023dit,esser2024sd3,lu2022dpmsolverpp}. Future work should test richer prompts, selective head aggregation~\cite{park2026selectiveaggregation}, additional schedulers, and newer backbones.

Participants may also overread the visualizations as proof of causal grounding. To reduce this risk, the instructions describe panels as attention-derived summaries and ask participants to cite maps, steps, or overlap regions. The current user study is a formative pilot with eight responses, useful for identifying communication strengths and design problems but not sufficient for statistical claims about broad usability. Looking forward, the same representation could support prompt clustering, live prompt editing, and human-in-the-loop controls for steering attention trajectories.

\section{Conclusion}
\label{sec:conclusion}
We presented a visual analytics framework for step-resolved diffusion attention. By linking DAAM-style token maps, phase summaries, and spatial competition views, the system helps readers reason about when attention changes and where token roles diverge while preserving a conservative distinction between attention-derived evidence and causal claims. These results support more transparent human--AI collaboration and point toward future interfaces where users can not only inspect attention dynamics, but also guide how attention develops during generation.

\FloatBarrier

\section*{Supplemental Materials and Online Demos}
\label{sec:supplemental_materials}

Supplemental materials include the prompt benchmark and prompt metadata, cached attention-derived summary tables, and anonymized study materials. We provide online demos as interactive Hugging Face Spaces: the paper-specific \href{https://huggingface.co/spaces/Shawyr/Attention-Dynamics-Explorer}{Attention Dynamics Explorer} uses precomputed data for step-resolved analysis, while the companion \href{https://huggingface.co/spaces/Shawyr/StepX}{StepX} interface supports live image generation and attention exploration.

\section*{Figure Credits}
\label{sec:figure_credits}

\Cref{fig:teaser,fig:daam_pipeline,fig:findings,fig:case_spatial_table} are original figures from the authors' pipeline; no external photography is used.

\acknowledgments{
This work was supported in part by a Visual, Performing, and Media Arts Award from the Interdisciplinary Humanities Center, University of California, Santa Barbara. We thank the pilot study participants for their feedback. AI-assisted tools were used for limited code debugging.}

\end{document}